\documentclass{article}
\usepackage{spconf,amsmath,graphicx}

\usepackage{amsmath,amssymb} 
\usepackage{color}
\usepackage{epsfig}
\usepackage{comment}
\usepackage{enumitem}
\usepackage[export]{adjustbox}
\usepackage{tabularx}
\usepackage[table]{xcolor}
\usepackage{colortbl}
\usepackage{soul}
\usepackage{multirow}
\usepackage{booktabs}
\usepackage{xr}
\usepackage{pifont}
\usepackage{hyperref}
\usepackage{sidecap}
\usepackage{ragged2e}
\usepackage{dsfont}

\title{GaitTAKE: Gait Recognition by Temporal Attention \\and Keypoint-guided Embedding}
\name{Hung-Min Hsu$^1$, Yizhou Wang$^1$, Cheng-Yen Yang$^1$, Jenq-Neng Hwang$^1$, \textit{Hoang Le Uyen Thuc}$^2$, \textit{Kwang-Ju Kim}$^3$}
\address{
$^1$ Information Processing Lab, University of Washington, Seattle, USA\\
$^2$ University of Science and Technology, University of Danang, Danang, Vietnam\\
$^3$ Electronics and Telecommunications Research Institute (ETRI), Daegu, South Korea}
\begin{document}
%
\maketitle
\begin{abstract}
Gait recognition, which refers to the recognition or identification of a person based on their body shape and walking styles, derived from video data captured from a distance, is widely used in crime prevention, forensic identification, and social security. However, to the best of our knowledge, most of the existing methods use appearance, posture and temporal feautures without considering a learned temporal attention mechanism for global and local information fusion. In this paper, we propose a novel gait recognition framework, called Temporal Attention and Keypoint-guided Embedding (GaitTAKE), which effectively fuses temporal-attention-based global and local appearance feature and temporal aggregated human pose feature. Experimental results show that our proposed method achieves a new SOTA in gait recognition with rank-1 accuracy of 98.0\% (normal), 97.5\% (bag) and 92.2\% (coat) on the CASIA-B gait dataset; 90.4\% accuracy on the OU-MVLP gait dataset.
\end{abstract}
\section{Introduction}
\label{sec:intro}


Gait recognition, which uses video data captured from a distance to recognize or identify a person based on their body shape and walking styles, is wildly used in crime prevention, forensic identification, and social security, etc.
Person re-identification (ReID) is one of the most popular research in the computer vision community. However, merely using the appearance feature is not sufficient to deal with some difficult scenarios, e.g., the same identity dressing different clothing, low resolution videos, the dark illumination cases. Therefore, gait recognition can serve as an effective supplement or alternative to overcome these issues.


There are two popular ways to recognize gaits in literatures, i.e., model-based \cite{wagg2004automated,yam2004automated,cunado2003automatic,ariyanto2012marionette} and appearance-based \cite{han2005individual,xu2006human,guan2014reducing,shiraga2016geinet}. The model-based approaches focus on the articulated human features such as the size of a link or joint angles, which can tolerate the appearance changes of an identity due to the clothings or accessories. These approaches require to preprocess the raw RGB videos to capture the pose structure or silhouettes. On the other hand, several studies have proposed appearance-based gait recognition approaches, which use RGB image sequences as input to recognize the identities directly. However, model-based approaches lose the body shape information and require high accuracy human pose estimation results for gait recognition. Moreover, appearance-based approaches suffer from the sensitivity to the identities’ covariates (e.g., dressing and carrying conditions).
In this paper, we propose a novel framework to generate the \textbf{T}emporal \textbf{A}ttention and \textbf{K}eypoint-guided \textbf{E}mbeddings in a principle way called GaitTAKE. The intuition of GaitTAKE is to take both global and local appearance features into account, then the learning of the silhouette embedding is trained by the temporal information. Thus, we can not only solve the flaws by the temporal pooling but also fuse the temporal information into the global and local features. Moreover, we combine the human pose information with the mentioned global and local features so that our method can achieve the large amount of improvement in the wearing coat scenario of gait recognition, which is the most difficult case in gait recognition since the coat will cover most of the area of the human legs. GaitTAKE forms embeddings over multiple frames with a global and local convolutional neural network \cite{lin2020learning} and human pose information with temporal attention mechanism. According to our experimental results, GaitTAKE achieves the state-of-the-art performance in CASIA-B \cite{yu2006framework} and OU-MVLP \cite{takemura2018multi} benchmarks.





\begin{figure*}[t]
    \centering
    \includegraphics[width=0.8\textwidth]{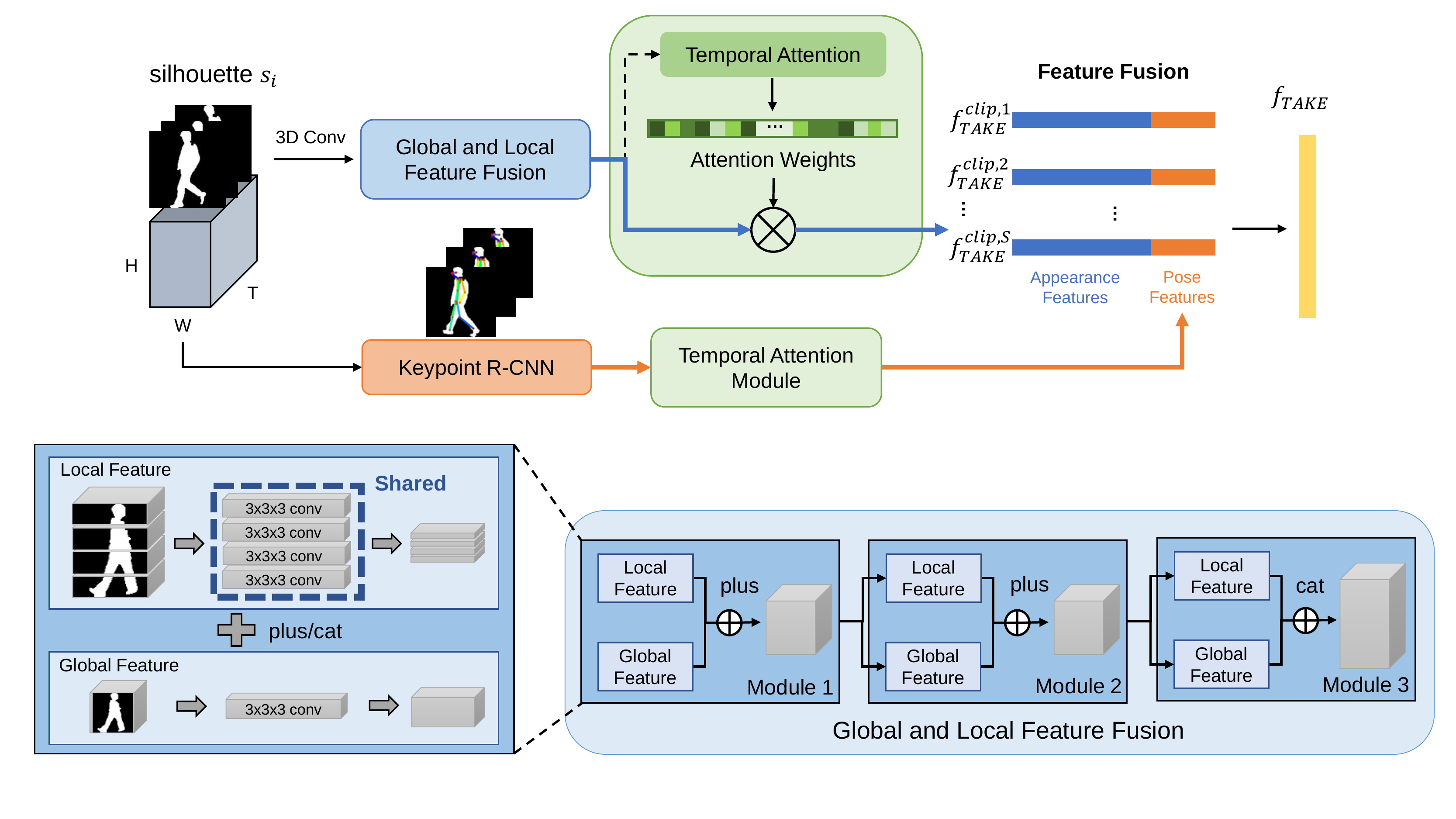}
    \caption{The architecture of the proposed GaitTAKE framework. First, the frame-level features for the gait videos are extracted by a global and local feature fusion backbone and a human pose feature extraction backbone Keypoint-RCNN. The extracted frame-level features (i.e., appearance features and human pose features) are fed into the temporal attention (TA) module. The TA weights are applied on the frame-level features to obtain the sequence-level features for each gait video.}
    \label{fig:framework}
\end{figure*}

\section{Related Works}
\label{sec:Related Works}

Due to the growth of deep learning, many researchers exploit convolutional neural networks (CNNs) to achieve great improvement for gait recognition \cite{shiraga2016geinet, wu2016comprehensive,chao2019gaitset,fan2020gaitpart,lin2020gait,li2020end}. The feature representation ability is robust, e.g., cross-view gait sequence can be recognized only based on the CNN feature and well-designed loss function. 


In terms of taking advantage of temporal information, there are two types of deep learning approaches: recurrent neural networks (RNNs) and 3D CNNs. In RNNs, the features are learned by a sequence of consecutive frames \cite{li2020end,liao2017pose,liu2016memory}. For a 3D CNN, the spatio-temporal information can be extracted by the 3D tensors \cite{lin2020gait,xing2018view,lin2020learning}. Nonetheless, there is a limitation of using 3D CNNs for gait recognition, which is the lack of flexibility for variable-length sequences. 

\section{Proposed Method}
\label{sec:Proposed Method}



\subsection{TA-based Global and Local Feature Fusion}
\label{sec:backbone}

As shown in Fig.~\ref{fig:framework}, the proposed feature extraction network architecture aims to simultaneously extract the global and local feature information and temporal information from the silhouette images. First, the 3D convolution is applied to extract more representative features from the silhouette images, since 3D convolution network is proved as an effective feature extractor for gait recognition \cite{lin2020learning, lin2020gait,xing2018view}. After that, the local information is extracted by horizontally dividing the image into several body partitions and the global information is extracted by a whole silhouette image. In order to aggregate the temporal information into the local and global feature maps, two 3D convolutions are applied to the local feature maps and the global feature map, separately. The local feature maps (i.e., body partition features) share the same weights of 3D convolutions. According to \cite{lin2020learning}, the generated global and local features can be added into one feature map to ensemble both global and local information. Then, this global and local feature fusion operation is repeated with the same network configuration and different convolution kernels for $n$ times to generate the more robust global and local fusion features.


The first step of generating the TA-based global and local feature fusion is to first generate the global and local features separately. We use $X \in \mathbb{R}^{c_1 \times T \times h \times w}$ to represent one sequence of silhouette with length $T$ (the image size is $h \times w$), and $\{X_{local}^i|i=1, \cdots ,m\}$ denotes the $m$ local gait partition features. $c$ is the channel size of the feature map. Thus, we can express the global gait feature ${f}_{global}$ as
\begin{equation}
{f}_{global}(X) = 
 \mathbf{\phi}_{global}^{3\times3\times3}(X) \in \mathbb{R}^{c_2 \times T \times h \times w},
\end{equation}
where $\mathbf{\phi}_{global}^{3\times3\times3}$ denotes 3D convolution operation with kernel size $3 \times 3 \times 3$. And for local gait feature ${f}_{local}$, similar mechanism is applied with shared 3D convolution kernels,
\begin{equation}
\begin{aligned}
{f}_{local}(X) &= {f}_{local}(\{X_{local}^i|i=1, \cdots ,m\}) \\
&= \mathbf{\phi}_{local}^{3\times3\times3}(X_{local}^1) \ \oplus \cdots \oplus \ \mathbf{\phi}_{local}^{3\times3\times3}(X_{local}^m) \\
&\in \mathbb{R}^{c_2 \times T \times h \times w},
\end{aligned}
\end{equation}
where $\phi_{local}^{3\times3\times3}$ is the shared 3D convolutional layer with kernel size $3 \times 3 \times 3$; $\oplus$ indicates the concatenation operation.


The TA fusion module is composed of two different structures of global and local convolutional (GLConv) layers, i.e., GLConvA and GLConvB. Fig.~\ref{fig:framework} shows that there are $n$ GLConv layers in this module for generating the global and local information fusion feature ${f}_{GL}$ ($n=3$). The last GLConv layer is GLConvB and the rest of other GLConv layers are GLConvA,
\begin{equation}
\begin{aligned}
{GLConvA}(X) &= f_{global}(X)  + f_{local}(X) \\
&\in \mathbb{R}^{c_2 \times T \times h \times w}.
\end{aligned}
\label{eq:dist}
\end{equation}
\begin{equation}
\begin{aligned}
{GLConvB}(X) &= f_{global}(X)  \oplus f_{local}(X) \\
&\in \mathbb{R}^{c_2 \times T \times 2h \times w}.
\end{aligned}
\end{equation}
Therefore, we can apply flatten operation $\xi(\cdot)$ to get the global and local information fused feature ${f}_{GL}$,
\begin{equation}
\begin{aligned}
{f}_{GL} &= \xi({GLConvB}({GLConvA}({GLConvA}(X)))) \\
&\in \mathbb{R}^{T \times D_{GL}},
\end{aligned}
\end{equation}
where $D_{GL}$ is the dimension of the ${f}_{GL}$.

After obtaining the global and local information fused feature ${f}_{GL}$, we can start to apply the TA mechanism to generate the final embedding ${f}_{TGL}$. First of all, the sequence of each subject is split into several clips. Assume the clip size is $L$, $S = \left \lfloor \frac{T}{L} \right \rfloor$ is the number of clips, and $D$ indicates the dimension of the clip-level feature.
\begin{equation}
{f}_{GL}^{clip} = \{f_{GL}^{clip,1}, \cdots , f_{GL}^{clip,S}\} \in \mathbb{R}^{S \times L \times D}.
\end{equation}
Then, there are two convolutional layers used for each clip in the TA module $\mathcal{T}_{GL}(\cdot)$ to produce a feature vector. We subsequently apply a softmax layer to this feature vector to generate a $1 \times L$-dim attention vector $\mathcal{A}_{GL}$ for weighting the frame-level feature so that the clip-level feature $f_{TGL}^{clip,i} \in \mathbb{R}^{1\times D}$ can be created.
\begin{equation}
\begin{aligned}
{f}_{TGL}^{clip,i} = \mathcal{T}_{GL}(f_{GL}^{clip,i})
= \mathcal{A}_{GL} \cdot f_{GL}^{clip,i}  \in \mathbb{R}^{1 \times D}.\\
\end{aligned}
\end{equation}
\begin{equation}
\mathcal{A}_{GL} = \sigma_{GL}(\delta_{GL,2}(\delta_{GL,1}(f^{clip,i}_{GL})))  \in \mathbb{R}^{1 \times L}.
\end{equation}
where $\sigma_{GL}(\cdot)$ is the softmax operation; $\delta_{GL,1}$ and $\delta_{GL,2}$ denote the first and second convolutional layer, respectively. 

Finally, one average pooling layer $\psi_{GL}(\cdot)$ is applied to these clip-level embeddings $f_{TGL}^{clip}$ to generate the final embedding $f_{TGL}$.
\begin{equation}
{f}_{TGL} = \psi_{GL}(f_{TGL}^{clip}) \in \mathbb{R}^{1 \times D}.
\end{equation}

\subsection{Temporal Aggregated Human Pose Feature}
\label{sec:pose}
In our framework, we not only consider the appearance embedding feature but also the human pose features since gait recognition is significantly related to the corresponding human pose. We use the keypoint R-CNN \cite{he2017mask} to obtain the human pose information. Since not all of the gait recognition datasets contain the human pose information, we use a pretrained model which is trained on COCO dataset to infer the human pose information based on the available RGB images as the ground truth human pose labels. Then, we use the human pose label to train the keypoint R-CNN based on the silhouette images so that we can use the trained keypoint R-CNN model to infer the human pose information on the silhouette images. 


After the human pose is estimated, we use the resulting 2D keypoints (body joints) as the extra features for the gait recognition. The dimension of the human pose features $\mathcal{K}$ for each frame is $17 \times 3$, where 17 is the number of joints, and 3 denotes the 2D joint coordinates $(x, y)$ and corresponding confidence score $c$. Similar to the appearance features, we also apply the temporal attention technique on the human pose features to aggregate the frame-level features into the clip-based human pose features, and then concatenate the Temporal Aggregated Human Pose Feature with $f_{TGL}$ as the final representation $f^{clip}_{TAKE}$ for gait recognition.

Consequently, we use a Generalized-Mean pooling (GeM) \cite{lin2020learning} to integrate the spatial information into the feature maps. GeM can effectively generate the more robust representation from the spatial information, Traditionally, researchers fuse the feature from average pooling and max pooling results by a weighted sum, on the other hand,  GeM can directly fuse these two different operations to form a feature map, with $p=1$ being equal to average pooling and $p=\infty$ being equal to max pooling,
\begin{equation}
\label{eq:gem}    
{f}_{GeM} = (\psi_{GeM}((f_{TAKE})^p))^\frac{1}{p},
\end{equation}
where $\psi_{GeM(\cdot)}$ is an average pooling operation.

\subsection{Loss Function}

The last step of feature extraction is to apply $C$ multiple different fully connected layers to the same $f_{GeM}$ to generate $C$ one-dimensional embedding $f$. Thus, each subject can be represented by $C$ different embeddings, and all the $f$ of the subject is used to calculate the loss independently. The loss function of our architecture is triplet loss, which is wildly used and proved to have superior performance in ReID tasks. 

The definition of the triplet loss function is as follows:
\begin{equation}
l_{triplet}(a) = \left[ m + \sum_{p \in P(a)} w_p D_{ap} - \sum_{n \in N(a)} w_n D_{an}\right] _+,
\end{equation}
where $m$ is the margin, $D_{ap}$ and $D_{an}$ indicates the distances between the anchor sample $a$ to form the positive instance and negative instance, respectively. Moreover, $w_{p}$ and $w_{n}$ mean the weights of positive and negative instances.




\begin{table*}
\begin{center}
\footnotesize
\begin{tabular}{ccccccccccccccc}
\hline
\multicolumn{1}{|c|}{\multirow{2}{*}{Setting}} & \multicolumn{1}{c}{\multirow{2}{*}{Probe}} & \multicolumn{1}{|c|}{\multirow{2}{*}{Method}} & \multicolumn{12}{c|}{Probe View} \\ 
\cline{4-15} 
\multicolumn{1}{|c|}{} &
\multicolumn{1}{c|}{} & \multicolumn{1}{c|}{} & \multicolumn{1}{c}{0$^{\circ}$}             & \multicolumn{1}{c}{18$^{\circ}$}            & \multicolumn{1}{c}{36$^{\circ}$}            & \multicolumn{1}{c}{54$^{\circ}$}            & \multicolumn{1}{c}{72$^{\circ}$}            & 
\multicolumn{1}{c}{90$^{\circ}$}   & \multicolumn{1}{c}{108$^{\circ}$}  & \multicolumn{1}{c}{126$^{\circ}$}  & \multicolumn{1}{c}{144$^{\circ}$}  & \multicolumn{1}{c}{162$^{\circ}$}  & \multicolumn{1}{c}{180$^{\circ}$}  & \multicolumn{1}{|c|}{\textbf{Mean}} \\ 
\hline\hline

\multicolumn{1}{|c|}{\multirow{15}{*}{LT(74)}} 
& \multicolumn{1}{c|}{\multirow{5}{*}{NM\#5-6}}    & \multicolumn{1}{c|}{Gaitset \cite{chao2019gaitset}}        & \multicolumn{1}{c}{90.8}          & \multicolumn{1}{c}{97.9}          & \multicolumn{1}{c}{\textbf{99.4}}          & \multicolumn{1}{c}{96.9}          & \multicolumn{1}{c}{93.6}          & \multicolumn{1}{c}{91.7} & \multicolumn{1}{c}{95.0} & \multicolumn{1}{c}{97.8} & \multicolumn{1}{c}{\underline{98.9}} & \multicolumn{1}{c}{96.8} & \multicolumn{1}{c}{85.8} & \multicolumn{1}{|c|}{95.0} \\ 
\multicolumn{1}{|c|}{} & \multicolumn{1}{c|}{}    & \multicolumn{1}{c|}{MT3D \cite{lin2020gait}}           & \multicolumn{1}{c}{\underline{95.7}}          & \multicolumn{1}{c}{98.2}          & \multicolumn{1}{c}{99.0}          & \multicolumn{1}{c}{97.5}          & \multicolumn{1}{c}{95.1}          & \multicolumn{1}{c}{93.9} & \multicolumn{1}{c}{96.1} & \multicolumn{1}{c}{\underline{98.6}} & \multicolumn{1}{c}{\textbf{99.2}} & \multicolumn{1}{c}{\underline{98.2}} & \multicolumn{1}{c}{\underline{92.0}} & \multicolumn{1}{|c|}{96.7} \\ 
\multicolumn{1}{|c|}{} & \multicolumn{1}{c|}{}    & \multicolumn{1}{c|}{GaitGL \cite{lin2020learning}}           & \multicolumn{1}{c}{94.6}          & \multicolumn{1}{c}{97.3}          & \multicolumn{1}{c}{{98.8}}          & \multicolumn{1}{c}{97.1}          & \multicolumn{1}{c}{\underline{95.8}}          & \multicolumn{1}{c}{\underline{94.3}} & \multicolumn{1}{c}{\underline{96.4}} & \multicolumn{1}{c}{98.5} & \multicolumn{1}{c}{98.6} & \multicolumn{1}{c}{\underline{98.2}} & \multicolumn{1}{c}{90.8} & \multicolumn{1}{|c|}{96.4} \\ 
\multicolumn{1}{|c|}{} & \multicolumn{1}{c|}{}    & \multicolumn{1}{c|}{GaitPart \cite{fan2020gaitpart}}           & \multicolumn{1}{c}{94.1}          & \multicolumn{1}{c}{\underline{98.6}}          & \multicolumn{1}{c}{\underline{99.3}}          & \multicolumn{1}{c}{\textbf{98.5}}          & \multicolumn{1}{c}{94.0}          & \multicolumn{1}{c}{92.3} & \multicolumn{1}{c}{95.9} & \multicolumn{1}{c}{98.4} & \multicolumn{1}{c}{\textbf{99.2}} & \multicolumn{1}{c}{97.8} & \multicolumn{1}{c}{90.4} & \multicolumn{1}{|c|}{96.2} \\ 
\multicolumn{1}{|c|}{} & \multicolumn{1}{c|}{}    & \multicolumn{1}{c|}{\textbf{GaitTAKE} (Ours)}        & \multicolumn{1}{c}{\textbf{96.7}}          & \multicolumn{1}{c}{\textbf{98.6}}          & \multicolumn{1}{c}{{99.1}}          & \multicolumn{1}{c}{\underline{98.1}}          & \multicolumn{1}{c}{\textbf{97.3}}          & \multicolumn{1}{c}{\textbf{96.3}} & \multicolumn{1}{c}{\textbf{98.0}} & \multicolumn{1}{c}{\textbf{98.9}} & \multicolumn{1}{c}{\textbf{99.2}} & \multicolumn{1}{c}{\textbf{99.2}} & \multicolumn{1}{c}{\textbf{96.4}} & \multicolumn{1}{|c|}{\textbf{98.0}} \\
\cline{2-15} 
\multicolumn{1}{|c|}{} & \multicolumn{1}{c|}{\multirow{5}{*}{BG\#1-2}}                       & \multicolumn{1}{c|}{Gaitset \cite{chao2019gaitset}}        & \multicolumn{1}{c}{83.8}          & \multicolumn{1}{c}{91.2}          & \multicolumn{1}{c}{91.8}          & \multicolumn{1}{c}{88.8}          & \multicolumn{1}{c}{83.3}          & \multicolumn{1}{c}{81.0} & \multicolumn{1}{c}{84.1} & \multicolumn{1}{c}{90.0} & \multicolumn{1}{c}{92.2} & \multicolumn{1}{c}{94.4} & \multicolumn{1}{c}{79.0} & \multicolumn{1}{|c|}{87.2} \\ 
\multicolumn{1}{|c|}{} & \multicolumn{1}{c|}{}    & \multicolumn{1}{c|}{MT3D \cite{lin2020gait}}           & \multicolumn{1}{c}{\underline{91.0}}          & \multicolumn{1}{c}{\underline{95.4}}          & \multicolumn{1}{c}{\underline{97.5}}          & \multicolumn{1}{c}{94.2}          & \multicolumn{1}{c}{\underline{92.3}}          & \multicolumn{1}{c}{86.9} & \multicolumn{1}{c}{\underline{91.2}} & \multicolumn{1}{c}{\underline{95.6}} & \multicolumn{1}{c}{\underline{97.3}} & \multicolumn{1}{c}{\underline{96.4}} & \multicolumn{1}{c}{86.6} & \multicolumn{1}{|c|}{93.0} \\
\multicolumn{1}{|c|}{} & \multicolumn{1}{c|}{}    & \multicolumn{1}{c|}{GaitGL \cite{lin2020learning}}            & \multicolumn{1}{c}{90.3}          & \multicolumn{1}{c}{94.7}          & \multicolumn{1}{c}{95.9}          & \multicolumn{1}{c}{94.0}          & \multicolumn{1}{c}{91.9}          & \multicolumn{1}{c}{86.5} & \multicolumn{1}{c}{90.5} & \multicolumn{1}{c}{95.5} & \multicolumn{1}{c}{97.2} & \multicolumn{1}{c}{96.3} & \multicolumn{1}{c}{\underline{87.1}} & \multicolumn{1}{|c|}{92.7} \\ 
\multicolumn{1}{|c|}{} & \multicolumn{1}{c|}{}    & \multicolumn{1}{c|}{GaitPart \cite{fan2020gaitpart}}           & \multicolumn{1}{c}{89.1}          & \multicolumn{1}{c}{94.8}          & \multicolumn{1}{c}{96.7}          & \multicolumn{1}{c}{\underline{95.1}}          & \multicolumn{1}{c}{88.3}          & \multicolumn{1}{c}{\underline{94.9}} & \multicolumn{1}{c}{89.0} & \multicolumn{1}{c}{93.5} & \multicolumn{1}{c}{96.1} & \multicolumn{1}{c}{93.8} & \multicolumn{1}{c}{85.8} & \multicolumn{1}{|c|}{91.5} \\ 
\multicolumn{1}{|c|}{} & \multicolumn{1}{c|}{}    & \multicolumn{1}{c|}{\textbf{GaitTAKE} (Ours)}        & \multicolumn{1}{c}{\textbf{96.7}}          & \multicolumn{1}{c}{\textbf{97.0}}          & \multicolumn{1}{c}{\textbf{97.9}}          & \multicolumn{1}{c}{\textbf{97.6}}          & \multicolumn{1}{c}{\textbf{97.9}}          & \multicolumn{1}{c}{\textbf{95.7}} & \multicolumn{1}{c}{\textbf{97.0}} & \multicolumn{1}{c}{\textbf{98.2}} & \multicolumn{1}{c}{\textbf{99.0}} & \multicolumn{1}{c}{\textbf{99.0}} & \multicolumn{1}{c}{\textbf{96.4}} & \multicolumn{1}{|c|}{\textbf{97.5}} \\ 
\cline{2-15} 
\multicolumn{1}{|c|}{} & \multicolumn{1}{c|}{\multirow{5}{*}{CL\#1-2}}                       & \multicolumn{1}{c|}{Gaitset \cite{chao2019gaitset}}        & \multicolumn{1}{c}{61.4} & \multicolumn{1}{c}{75.4} & \multicolumn{1}{c}{80.7} & \multicolumn{1}{c}{77.3} & \multicolumn{1}{c}{72.1} & \multicolumn{1}{c}{70.1} & \multicolumn{1}{c}{71.5} & \multicolumn{1}{c}{73.5} & \multicolumn{1}{c}{73.5} & \multicolumn{1}{c}{68.4} & \multicolumn{1}{c}{50.0} & \multicolumn{1}{|c|}{70.4} \\ 
\multicolumn{1}{|c|}{} & \multicolumn{1}{c|}{}    & \multicolumn{1}{c|}{MT3D \cite{lin2020gait}}           & \multicolumn{1}{c}{76.0}          & \multicolumn{1}{c}{{87.6}}          & \multicolumn{1}{c}{{89.8}}          & \multicolumn{1}{c}{85.0}          & \multicolumn{1}{c}{81.2}          & \multicolumn{1}{c}{75.7} & \multicolumn{1}{c}{81.0} & \multicolumn{1}{c}{84.5} & \multicolumn{1}{c}{{85.4}} & \multicolumn{1}{c}{82.2} & \multicolumn{1}{c}{68.1} & \multicolumn{1}{|c|}{81.5} \\ 
\multicolumn{1}{|c|}{} & \multicolumn{1}{c|}{}    & \multicolumn{1}{c|}{GaitGL \cite{lin2020learning}}           & \multicolumn{1}{c}{\underline{76.7}} & \multicolumn{1}{c}{\underline{88.3}} & \multicolumn{1}{c}{\underline{90.7}} & \multicolumn{1}{c}{\underline{86.6}} & \multicolumn{1}{c}{\underline{82.7}} & \multicolumn{1}{c}{\underline{77.6}} & \multicolumn{1}{c}{\underline{83.5}} & \multicolumn{1}{c}{\underline{86.5}} & \multicolumn{1}{c}{\underline{88.1}} & \multicolumn{1}{c}{\underline{83.2}} & \multicolumn{1}{c}{\underline{68.7}} & \multicolumn{1}{|c|}{83.0} \\ 
\multicolumn{1}{|c|}{} & \multicolumn{1}{c|}{}    & \multicolumn{1}{c|}{GaitPart \cite{fan2020gaitpart}}           & \multicolumn{1}{c}{70.7}          & \multicolumn{1}{c}{85.5}          & \multicolumn{1}{c}{86.9}          & \multicolumn{1}{c}{83.3}          & \multicolumn{1}{c}{77.1}          & \multicolumn{1}{c}{72.5} & \multicolumn{1}{c}{76.9} & \multicolumn{1}{c}{82.2} & \multicolumn{1}{c}{83.8} & \multicolumn{1}{c}{80.2} & \multicolumn{1}{c}{66.5} & \multicolumn{1}{|c|}{78.7} \\ 
\multicolumn{1}{|c|}{} & \multicolumn{1}{c|}{}    & \multicolumn{1}{c|}{\textbf{GaitTAKE} (Ours)}       & \multicolumn{1}{c}{\textbf{89.1}}          & \multicolumn{1}{c}{\textbf{95.3}}          & \multicolumn{1}{c}{\textbf{96.2}}          & \multicolumn{1}{c}{\textbf{93.9}}          & \multicolumn{1}{c}{\textbf{91.5}}          & \multicolumn{1}{c}{\textbf{90.5}} & \multicolumn{1}{c}{\textbf{92.5}} & \multicolumn{1}{c}{\textbf{93.3}} & \multicolumn{1}{c}{\textbf{93.0}} & \multicolumn{1}{c}{\textbf{91.8}} & \multicolumn{1}{c}{\textbf{87.0}} & \multicolumn{1}{|c|}{\textbf{92.2}} \\ 
\hline
\end{tabular}
\caption{Rank-1 accuracy (\%) of the proposed method on CASIA-B under all views, diffeent size of training data and conditions, excluding identical-view cases. The  three walking conditions of sequences include normal (NM),  walking  with  bag (BG) and wearing coat or jacket (CL). The \textbf{best} and \underline{second} accuracy of each probe view will be in bold and underlined respectively.}\label{tab:casiab_results}
\end{center}
\end{table*}

\begin{table*}
\begin{center}
\footnotesize

\begin{tabular}{cccccccccccccccc}
\hline
\multicolumn{1}{|c|}{\multirow{2}{*}{Method}} & \multicolumn{15}{c|}{Probe View} \\ 
\cline{2-16} 
\multicolumn{1}{|c|}{} &
\multicolumn{1}{c}{0$^{\circ}$}    &      \multicolumn{1}{c}{15$^{\circ}$}   & \multicolumn{1}{c}{30$^{\circ}$}   & \multicolumn{1}{c}{45$^{\circ}$}   & \multicolumn{1}{c}{60$^{\circ}$}   & \multicolumn{1}{c}{75$^{\circ}$}   & \multicolumn{1}{c}{90$^{\circ}$}   & \multicolumn{1}{c}{180$^{\circ}$}  & \multicolumn{1}{c}{195$^{\circ}$}  & \multicolumn{1}{c}{210$^{\circ}$}  & \multicolumn{1}{c}{225$^{\circ}$}  & \multicolumn{1}{c}{240$^{\circ}$}  & \multicolumn{1}{c}{255$^{\circ}$}  & \multicolumn{1}{c}{270$^{\circ}$}  & \multicolumn{1}{|c|}{\textbf{Mean}} \\ 
\hline\hline
\multicolumn{1}{|c|}{Gaitset \cite{chao2019gaitset}} & \multicolumn{1}{c}{79.5} & \multicolumn{1}{c}{87.9} & \multicolumn{1}{c}{89.9} & \multicolumn{1}{c}{90.2} & \multicolumn{1}{c}{88.1} & \multicolumn{1}{c}{88.7}       & \multicolumn{1}{c}{87.8} & \multicolumn{1}{c}{81.7} & \multicolumn{1}{c}{86.7} & \multicolumn{1}{c}{89.0} & \multicolumn{1}{c}{89.3} & \multicolumn{1}{c}{87.2}       & \multicolumn{1}{c}{87.8} & \multicolumn{1}{c}{86.2} &  \multicolumn{1}{|c|}{87.1} \\ 
\multicolumn{1}{|c|}{GaitPart \cite{fan2020gaitpart}} & \multicolumn{1}{c}{82.6} & \multicolumn{1}{c}{88.9} & \multicolumn{1}{c}{\underline{90.8}} & \multicolumn{1}{c}{\underline{91.0}}       & \multicolumn{1}{c}{89.7} & \multicolumn{1}{c}{89.9} & \multicolumn{1}{c}{89.5} & \multicolumn{1}{c}{85.2} & \multicolumn{1}{c}{\underline{88.1}} & \multicolumn{1}{c}{\underline{90.0}}       & \multicolumn{1}{c}{\underline{90.1}} & \multicolumn{1}{c}{\underline{89.0}} & \multicolumn{1}{c}{\underline{89.1}} & \multicolumn{1}{c}{\underline{88.2}} &  \multicolumn{1}{|c|}{88.7} \\ 
\multicolumn{1}{|c|}{GaitGL \cite{lin2020learning}} & \multicolumn{1}{c}{\underline{84.3}}          & \multicolumn{1}{c}{\underline{89.8}} & \multicolumn{1}{c}{\underline{90.8}} & \multicolumn{1}{c}{\underline{91.0}}       & \multicolumn{1}{c}{\underline{90.5}} & \multicolumn{1}{c}{\underline{90.3}} & \multicolumn{1}{c}{\underline{89.9}}       & \multicolumn{1}{c}{\underline{88.1}} & \multicolumn{1}{c}{87.9} & \multicolumn{1}{c}{89.6}       & \multicolumn{1}{c}{89.8} & \multicolumn{1}{c}{88.9} & \multicolumn{1}{c}{88.9}       & \multicolumn{1}{c}{\underline{88.2}} &  \multicolumn{1}{|c|}{89.1} \\ 
\multicolumn{1}{|c|}{\textbf{GaitTAKE} (Ours)} & \multicolumn{1}{c}{\textbf{87.5}}       & \multicolumn{1}{c}{\textbf{91.0}}       & \multicolumn{1}{c}{\textbf{91.5}}       & \multicolumn{1}{c}{\textbf{91.8}}       & \multicolumn{1}{c}{\textbf{91.4}}       & \multicolumn{1}{c}{\textbf{91.1}}       & \multicolumn{1}{c}{\textbf{90.8}}       & \multicolumn{1}{c}{\textbf{90.2}}       & \multicolumn{1}{c}{\textbf{89.7}}       & \multicolumn{1}{c}{\textbf{90.5}}       & \multicolumn{1}{c}{\textbf{90.7}}       & \multicolumn{1}{c}{\textbf{90.3}}       & \multicolumn{1}{c}{\textbf{90.0}}       & \multicolumn{1}{c}{\textbf{89.5}}       &  \multicolumn{1}{|c|}{\textbf{90.4}} \\ 

\hline

\end{tabular}
\caption{Rank-1 accuracy (\%) of the proposed method on OU-MVLP under 14 probe views excluding identical-view cases.}\label{tab:oumvlp_results}
\end{center}
\end{table*}

\section{Experiments}
\label{sec:Evaluation}

In this work, we use two benchmarks for evaluating the proposed GaitTAKE, namely CASIA-B and OU-MVLP. The first part of this section is to describe the details of the implementation. The second part is to compare GaitTAKE with other state-of-the-art methods in these two datasets.

\subsection{Implementation Details}

In our implementation, the batch size $P \times K$ is set to $8 \times 8$ = 64 in both CASIA-B and OU-MVLP datasets. Following \cite{chao2019gaitset}, we use 30 frames of each input gait sequence for training and the whole gait sequences are used for extracting gait features in testing. In terms of the number of GLConv layers $n$, we use 3 GLConv layers (i.e., GLConvA, GLConvA and GLConvB) for CASIA-B dataset. Because OU-MVLP dataset is 20 times larger than the CASIA-B dataset, we use a total of 5 layers, which are 4 GLConvA following by 1 GLConvB layer. 

Since the CASIA-B dataset does not contain the keypoint information, we adopt the pre-trained Keypoint R-CNN trained on COCO to estimate the keypoints, which are then used as ground-truth to train the Keypoint R-CNN using masked images, instead of RGB images, as input data. 
The experimental environment is Python 3.7 and Pytorch 1.7 with one Nivida GV100.



\subsection{Gait Recognition Performance}
\textbf{Evaluation on CASIA-B.}  We compare our method with state-of-the-art methods: Multiple-Temporal-Scale 3D Convolutional Neural Network (MT3D) \cite{lin2020gait}, Gaitset \cite{chao2019gaitset}, GaitPart \cite{fan2020gaitpart} and GaitGL \cite{lin2020learning} in three different conditions (NM, BG, and CL). 
The experimental results show that the rank-1 accuracy of the proposed method is higher than GaitGL by about 1.6\% and 4.8\% in NM and BG, and about 9.2\% in CL with the setting of large-scale training (LT, i.e., 74 subjects for training), respectively. It shows that the proposed method has significant advantages in the BG and CL conditions, indicating that the representation of GaitTAKE is much more discriminative than other state-of-the-of methods. 


\noindent\textbf{Evaluation on OU-MVLP.} We also evaluate the performance of GaitTAKE on the OU-MVLP dataset, where we follow the same training and test protocols as the GaitSet, GaitPart and GaitGL methods for fair comparison. 
The experimental results are shown in Table \ref{tab:oumvlp_results} and then our method can also achieve the best performance in all cases.

\section{Conclusion}
\label{sec:CONCLUSIONS}
In this paper, we propose GaitTAKE, which utilizes the temporal attention module to generate the embedding for multi-view gait recognition. We use human pose information and temporal attention to construct the more robust features. Our experimental results show that we can achieve the state-of-the-art performance rank-1 accuracy on two representative gait recognition benchmarks: CASIA-B and OU-MVLP dataset. 

\bibliographystyle{IEEEbib}
\bibliography{strings,refs}

\end{document}